\documentclass[runningheads]{llncs}

\usepackage{eccv}
\usepackage{eccvabbrv}
\usepackage[accsupp]{axessibility}
\usepackage{graphicx}
\usepackage{amsmath,amssymb}
\usepackage{siunitx}
\usepackage{xcolor}
\usepackage[most]{tcolorbox}
\usepackage{microtype}
\usepackage[breaklinks,colorlinks]{hyperref}
\usepackage{orcidlink}

\graphicspath{{figures/}}
\newcommand{\ape}{\mathrm{APE}}
\newcommand{\rpe}{\mathrm{RPE}}

\definecolor{summaryfill}{RGB}{244,249,252}
\definecolor{summaryframe}{RGB}{170,205,224}
\newtcolorbox{resultsummary}{
  enhanced,
  colback=summaryfill,
  colframe=summaryframe,
  boxrule=0.35pt,
  arc=1.4mm,
  outer arc=1.4mm,
  left=2.0mm,
  right=2.0mm,
  top=1.35mm,
  bottom=1.35mm,
  boxsep=0pt,
  before skip=6pt,
  after skip=7pt,
  fontupper=\small
}
\newcommand{\resultbox}[1]{\begin{resultsummary}#1\end{resultsummary}}

\begin{document}

\title{Failure or Drift? Evaluating Monocular SLAM under Synthetic and Real-World Corruptions}
\titlerunning{Failure or Drift? SLAM under Corruptions}

\author{Abhay Skaria Thomas\inst{1}\orcidlink{0009-0009-8004-7872} \and
Shashank Agnihotri\inst{1}\orcidlink{0000-0001-6097-8551} \and
Margret Keuper\inst{1,2}\orcidlink{0000-0002-8437-7993}}

\authorrunning{A.\ S.\ Thomas et al.}

\institute{Machine Learning Group, University of Mannheim, Germany \and
Max-Planck-Institute for Informatics, Saarland Informatics Campus, Germany
\\
\email{\{shashank.agnihotri,keuper\}@uni-mannheim.de}}

\maketitle

\begin{abstract}
Visual SLAM is commonly evaluated on clean trajectories, although deployment failures are often caused by adverse weather, illumination, blur, and sensor artifacts. Controlled corruptions are attractive because they isolate such factors, but a synthetic stress test is useful only when it leads to the same engineering conclusion as the condition it is intended to approximate. This work examines that question for monocular SLAM. We evaluate a classical feature-based system and two learned trackers under image-space, geometry-aware, and compound corruptions, and compare their behavior with adverse conditions from 4Seasons. Rather than reducing robustness to a single trajectory error, the evaluation separates explicit tracking failure from drift accumulated by methods that remain active. The results show that learned trackers largely replace catastrophic loss with sustained, and sometimes severe, drift. More importantly, the apparent ordering of the learned systems changes with the physical fidelity of the corruption: structured rain and fog proxies preserve the real-world ordering, whereas a simple illumination proxy does not. Code is available in this: \hyperlink{https://github.com/abhaythomas/master_thesis_vslamlab_robustness}{GitHub repository}. 
\end{abstract}

\section{Introduction}
\label{sec:introduction}

Camera tracking is a prerequisite for autonomous driving, mobile robotics, augmented reality, and online 3D reconstruction. The standard evaluation of a monocular SLAM system nevertheless remains dominated by clean benchmark trajectories. Such measurements are necessary, but they do not describe the conditions under which a system will be used. Cameras encounter rain, fog, low illumination, motion blur, compression, exposure variation, and sensor noise~\cite{muller2023classification,sommerhoff2024task,stracke2025vision,fatima2025gamma,fatima2026rawdet}. These changes weaken visual correspondences and can disrupt both local tracking and the optimization that carries information across time.

The resulting failure is not always explicit. A feature-based system can lose tracking and return no usable trajectory. A learned tracker can remain active while its error accumulates over the sequence. The first outcome is easy to detect; the second can appear well-behaved until an incorrect pose reaches mapping, planning, or control. Robustness must therefore distinguish whether a trajectory is available from whether the available trajectory remains accurate.

Real adverse-condition datasets provide the most faithful evidence, but they make controlled diagnosis difficult. In 4Seasons~\cite{wenzel2021fourseasons}, for example, weather and illumination change together with route geometry, traffic, camera response, and motion. Synthetic corruptions provide the complementary intervention: the route, calibration, timestamps, and ground truth can remain fixed while the visual input changes. This control has made common-corruption benchmarks useful across vision tasks. For a stateful system such as SLAM, however, visual plausibility is not sufficient. A corruption may look similar to an adverse condition while changing a different subset of correspondences and, consequently, favoring a different tracker.

This work studies the external validity of such stress tests. We compare ORB-SLAM2~\cite{murartal2017orbslam2}, DPVO~\cite{teed2023dpvo}, and DROID-SLAM~\cite{teed2021droid} without corruption-specific adaptation. Clean performance is established on KITTI~\cite{geiger2012kitti} sequences 00--10. The controlled study uses 115 corrupted variants of KITTI sequence 00, covering image-space, geometry-aware, and compound effects at five severity levels. Four 4Seasons~\cite{wenzel2021fourseasons} sequences provide reference, rain, winter or fog-like, and evening conditions for a conservative synthetic-to-real comparison.

The findings expose two independent decisions. First, ORB-SLAM2 often fails by returning no evaluable trajectory, whereas both learned systems continue in every archived synthetic attempt and carry the risk of substantial drift. Second, the corruption model changes the comparison between the learned trackers. DPVO has lower APE in 30 of 43 paired image-space settings, while DROID-SLAM has lower APE in all 33 paired geometry-aware settings and 14 of 15 compound settings. Rain and winter preserve the compound-proxy ordering in real data; evening illumination reverses it. A benchmark constructed only from convenient image transformations can therefore recommend the wrong architecture for the condition it is intended to represent.

Our main contributions are:
\begin{itemize}
  \item a controlled evaluation of three monocular SLAM paradigms across image-space, geometry-aware, compound, and natural adverse conditions;
  \item a failure-aware protocol that reports trajectory validity before APE and RPE conditioned on a valid output, avoiding comparisons that conflate tracking loss with exporter density;
  \item evidence that corruption fidelity changes tracker selection, together with a direct analysis of which synthetic conclusions transfer to real adverse data.
\end{itemize}

\section{Related Work}
\label{sec:related}
In the following, we review the monocular SLAM paradigms and robustness benchmarks most relevant to this work, before discussing synthetic corruption protocols and their validity as proxies for real adverse conditions.
\subsection{Monocular Tracking and Dense Scene Representations}
ORB-SLAM2~\cite{murartal2017orbslam2} combines ORB~\cite{rublee2011orb} keypoints with geometric pose estimation, local mapping, loop closure, and relocalization. Its explicit front end provides interpretable failure signals, but initialization and tracking depend on repeatable local descriptors. DPVO~\cite{teed2023dpvo} replaces handcrafted keypoints with learned patch correspondences and couples a recurrent update operator with differentiable bundle adjustment. DROID-SLAM~\cite{teed2021droid} uses dense learned correspondences and repeatedly refines poses and depth through dense bundle adjustment. These systems span the three behaviors central to this study: discrete correspondence failure, continued local odometry with accumulated drift, and learned tracking supported by global multi-frame optimization.

Recent systems couple tracking more tightly to dense reconstruction. GO-SLAM~\cite{zhang2023goslam} jointly optimizes camera poses and an implicit scene representation. Gaussian Splatting SLAM~\cite{matsuki2024gaussian}, SplaTAM~\cite{keetha2024splatam}, and GS-SLAM~\cite{yan2024gsslam} use 3D Gaussians for tracking and mapping. MASt3R-SLAM~\cite{murai2025mast3r} builds a real-time dense pipeline around a two-view reconstruction prior, while SLAM3R~\cite{liu2025slam3r} directly regresses and registers local pointmaps. Their progress makes adverse-condition evaluation increasingly important: an appearance shift can affect not only pose estimation, but also geometry, rendering, map updates, and memory.

\subsection{Robustness Evaluation for SLAM}
KITTI~\cite{geiger2012kitti} and the TUM RGB-D benchmark~\cite{sturm2012tum} established reproducible trajectory evaluation under nominal conditions. TartanAir~\cite{wang2020tartanair} broadened the range of simulated environments, motion, and weather, while 4Seasons~\cite{wenzel2021fourseasons} introduced repeated outdoor recordings across seasons, illumination, and weather. Such datasets reveal realistic failures, but adverse conditions remain entangled with route and sensor differences~\cite{prokhorov2019measuring,agnihotri2024cospgd,agnihotri2025flowbench,hoffmann2021towards}.

Measuring Robustness of Visual SLAM~\cite{prokhorov2019measuring} demonstrated the importance of evaluating variation across more than a small set of trajectories. VSLAM-LAB~\cite{fontan2025vslamlab,vslamlabgithub} addresses a complementary source of variation by standardizing compilation, dataset preparation, configuration, execution, and evaluation. SLAM Adversarial Lab~\cite{hefny2026sal} provides modular perturbations, interpretable severity units, and a search for failure boundaries. Here, we ask a different question: does the conclusion obtained from a controlled perturbation remain stable when the proxy becomes more physically structured and is then tested on natural adverse data?

\subsection{Synthetic Corruptions and Their Real-World Validity}
ImageNet-C~\cite{hendrycks2019common} established parameterized common corruptions as a practical robustness protocol. The imagecorruptions package~\cite{michaelis2019winter,imagecorruptionsgithub} exposes many of these transformations for arbitrary images. 3D Common Corruptions~\cite{kar2022common} incorporates depth and scene geometry into effects such as fog and focus changes. RobustSpring~\cite{oei2026robustspring} extends corruption evaluation to optical flow, scene flow, and stereo with temporal, stereo, and depth consistency.

Are Synthetic Corruptions A Reliable Proxy For Real-World Corruptions?~\cite{agnihotri2025synthetic,agnihotri2025flowbench,agnihotri2025semsegbench} shows that strong aggregate correspondence in semantic segmentation can coexist with corruption-specific mismatch. The question is more demanding for SLAM because errors propagate through a stateful estimation process. A local correspondence change can alter subsequent poses, optimization, and map state. We therefore assess synthetic-to-real validity through failure behavior and relative method ordering, not through equality of absolute trajectory errors.

\section{Experimental Design}
\label{sec:design}
To isolate the effect of visual degradation, all methods are evaluated under a common protocol without corruption-specific adaptation. The following subsections describe the datasets and corruptions, evaluated systems, and failure-aware trajectory evaluation.
\subsection{Datasets and Corruptions}
The clean baseline contains the eleven KITTI~\cite{geiger2012kitti} odometry training sequences with ground-truth trajectories, evaluated in monocular mode. The controlled corruption study is generated from sequence 00. This sequence is sufficiently long to expose both local tracking errors and long-range drift, while keeping route, calibration, timestamps, and ground truth fixed across conditions. The use of a single controlled route is intentional for diagnosis and is treated as a limitation for generalization.

The image-space branch contains 13 transformations from the imagecorruptions package~\cite{michaelis2019winter,imagecorruptionsgithub}: brightness, contrast, defocus blur, elastic transform, frost, Gaussian noise, impulse noise, JPEG compression, motion blur, pixelation, shot noise, snow, and zoom blur. The geometry-aware branch follows 3D Common Corruptions~\cite{kar2022common} and contains far focus, near focus, flash, fog, ISO noise, low light, and rain. These effects use depth or physically motivated scene structure where applicable. The compound branch combines brightness with contrast, fog with contrast, and rain with contrast. Each family has five nominal severity levels. The complete synthetic collection therefore contains 65 image-space, 35 geometry-aware, and 15 compound sequence variants.

The real-world comparison uses four neighborhood sequences from 4Seasons~\cite{wenzel2021fourseasons}. We use neighborhood 2 train as a cloudy-afternoon reference, neighborhood 3 train as rainy afternoon, neighborhood 4 train as winter cloudy morning, and neighborhood 6 train as cloudy evening. The corresponding synthetic conditions are clean KITTI, rain with contrast, fog with contrast, and brightness with contrast. These are condition-level matches rather than paired recordings; route, camera, motion, and scene geometry remain different.

\subsection{Systems and Execution}
ORB-SLAM2~\cite{murartal2017orbslam2} represents sparse feature-based SLAM with loop closure and relocalization. DPVO~\cite{teed2023dpvo} represents learned patch-based visual odometry without classical loop closure. DROID-SLAM~\cite{teed2021droid} represents dense learned correspondence with global multi-frame optimization. Released weights and dataset-appropriate settings are used without fine-tuning on corrupted data, test-time restoration, or condition-specific adaptation.

All runs are executed through VSLAM-LAB~\cite{fontan2025vslamlab,vslamlabgithub}. Dataset and experiment YAML files define image paths, timestamps, calibration, method modules, and method-specific parameters. Every execution archives its trajectory, logs, runtime records, and evaluation outputs. ORB-SLAM2 is executed on CPU; DPVO and DROID-SLAM use NVIDIA A100 or H100 GPUs. The archive contains more than 350 executions. Missing configurations are retained as missing and are never converted into failures or imputed measurements. Quantitative comparisons between DPVO and DROID-SLAM use only conditions for which both results are present.

\subsection{Failure-Aware Evaluation}
For an attempted run $r$, we first record whether it returns a non-empty trajectory that can be associated with ground truth,
\begin{equation}
V_r = \begin{cases}
1, & \text{evaluable trajectory returned},\\
0, & \text{otherwise}.
\end{cases}
\end{equation}
The valid-run rate is the mean of $V_r$ over archived attempts. This binary measure is deliberately narrower than trajectory coverage. ORB-SLAM2 commonly exports keyframes, whereas the learned systems can export poses at a different temporal density. Some archived corruption sequences are also subsampled. Dividing the number of exported poses by the original number of frames would therefore mix keyframe density, input sampling, and genuine tracking loss.

For $V_r=1$, poses are associated within \SI{0.01}{s} and the monocular trajectory is aligned by a similarity transform $S\in\mathrm{Sim}(3)$. We report translation RMSE for the standard absolute and relative pose errors used by the TUM RGB-D benchmark~\cite{sturm2012tum},
\begin{align}
\ape_{\mathrm{RMSE}} &= \sqrt{\frac{1}{N}\sum_{i=1}^{N}
\left\|\operatorname{trans}\!\left(T_i^{-1}S\hat T_i\right)\right\|_2^2},\\
\rpe_{\mathrm{RMSE}} &= \sqrt{\frac{1}{M}\sum_{i=1}^{M}
\left\|\operatorname{trans}\!\left((T_i^{-1}T_{i+\Delta})^{-1}
(\hat T_i^{-1}\hat T_{i+\Delta})\right)\right\|_2^2}.
\end{align}
APE measures accumulated global disagreement; RPE measures local motion inconsistency. Both are conditioned on a valid trajectory so that a method cannot obtain a favorable error by returning no output.

To compare the learned trackers, we use
\begin{equation}
\Delta\ape = \ape_{\mathrm{DPVO}}-\ape_{\mathrm{DROID\text{-}SLAM}},
\end{equation}
where a positive value favors DROID-SLAM. For synthetic-to-real analysis, each compound result is averaged over its five severity levels before it is compared with the corresponding real condition. We also report descriptive Spearman correlations~\cite{spearman1904} between geometry-aware and compound severity curves. Since each correlation contains only five severity points, it is used as a diagnostic rather than an inferential statistic.

\section{Results}
\label{sec:results}
We organize the results around four questions concerning how the systems fail, whether corruption fidelity changes tracker selection, how conclusions vary across proxy layers, and which findings transfer to real adverse conditions.
\subsection{Do the Systems Fail, or Do They Drift?}
The clean KITTI runs provide an important reference before corruption is introduced. ORB-SLAM2, DPVO, and DROID-SLAM each return an evaluable trajectory on all 11 sequences. ORB-SLAM2 exports a sparser keyframe trajectory, so its raw number of stored poses is not interpreted as cross-method coverage. Yet clean input does not produce a stable accuracy ordering. ORB-SLAM2 has the lowest median conditional APE (\SI{8.2}{m}), but reaches \SI{518.0}{m} on sequence 01. DPVO and DROID-SLAM have median APEs of \SI{64.9}{m} and \SI{82.0}{m}, respectively, and their relative performance changes with the route: DPVO is substantially better on sequence 01, whereas DROID-SLAM is better on sequence 08. Even without corruption, successful execution and dependable global accuracy are therefore different properties.

\begin{figure}[t]
  \centering
  \includegraphics[width=\textwidth]{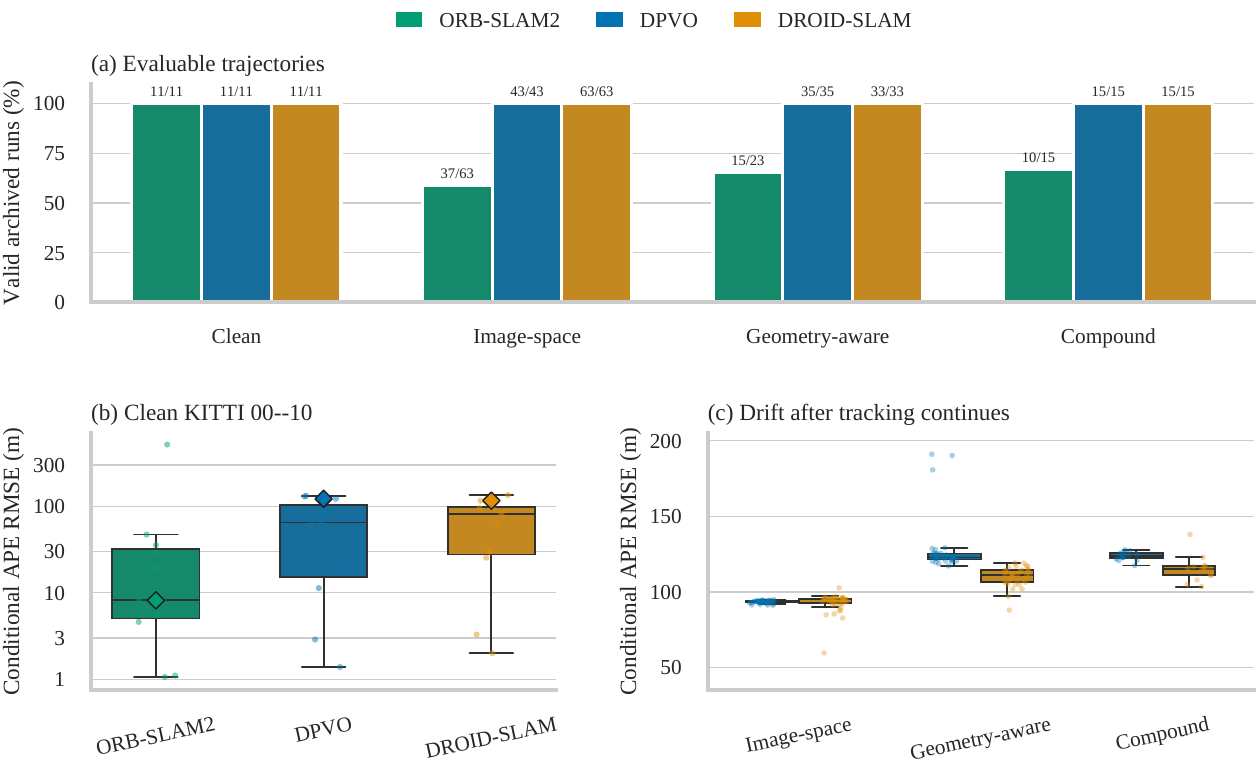}
  \caption{Clean and corrupted inputs reveal different aspects of failure. (a) All three systems return an evaluable trajectory on each clean KITTI sequence; under synthetic corruption, ORB-SLAM2 frequently returns no evaluable trajectory, while DPVO and DROID-SLAM remain valid in every archived attempt. Missing, unexecuted configurations are excluded from the denominators. (b) Conditional APE across clean KITTI sequences 00--10; diamonds mark sequence 00, the nominal reference used for the corruption sweep. The clean distribution shows strong route dependence and is not a matched degradation curve for every corrupted archive. (c) Conditional APE for paired DPVO and DROID-SLAM settings after tracking continues.}
  \label{fig:failure}
\end{figure}

Figure~\ref{fig:failure} shows how corruption changes this scenario. DPVO and DROID-SLAM return an evaluable trajectory for every archived synthetic attempt. ORB-SLAM2 is valid in 37 of 63 image-space attempts, 15 of 23 geometry-aware attempts, and 10 of 15 compound attempts. The image-space archive contains 25 empty ORB-SLAM2 trajectories and one synchronization failure. Gaussian noise, impulse noise, and snow produce no evaluable ORB-SLAM2 trajectory at any recorded severity; brightness and JPEG compression are substantially less destructive.

The corruption results expose a sharper difference between explicit failure and silent degradation. ORB-SLAM2 often stops returning an evaluable trajectory, whereas the learned systems continue to produce poses while their aligned APE frequently exceeds \SI{100}{m} under geometry-aware and compound effects. This distinction matters operationally. Loss of tracking can trigger relocalization or sensor fallback; a continuous but inaccurate estimate may pass through the rest of the autonomy stack without a comparable warning.

\resultbox{\textbf{Clean data reveal route-dependent drift; corruption changes the failure mechanism.} All three systems return evaluable trajectories on clean KITTI, although their errors vary strongly by route. Under corruption, ORB-SLAM2 often fails explicitly, whereas DPVO and DROID-SLAM remain active and can accumulate substantial drift.}

\subsection{Does Corruption Fidelity Change Which Tracker Appears Robust?}
It does, and the reversal is systematic. Among the 43 paired image-space settings with complete learned-method results, DPVO has lower APE in 30. Brightness and contrast favor DPVO at every tested severity. A benchmark restricted to these inexpensive global transformations would therefore select DPVO.

\begin{figure}[t]
  \centering
  \includegraphics[width=0.94\textwidth]{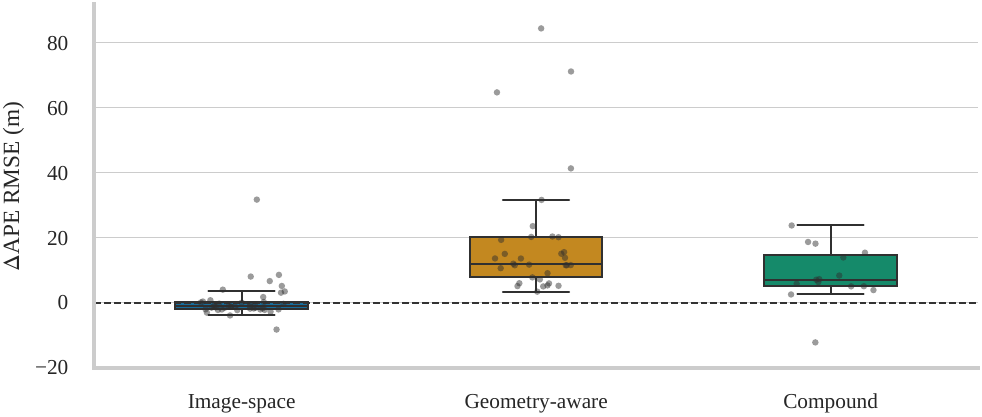}
  \caption{The corruption model changes tracker selection. Each point is a paired condition with complete DPVO and DROID-SLAM results; $\Delta\ape>0$ favors DROID-SLAM. DPVO has lower APE in 30/43 image-space settings, whereas DROID-SLAM has lower APE in 33/33 geometry-aware settings and 14/15 compound settings.}
  \label{fig:selection}
\end{figure}

The geometry-aware branch produces the opposite result. DROID-SLAM has lower APE in all 33 paired settings, including every available severity of low light, fog, rain, flash, ISO noise, and focus changes. The compound branch largely preserves that ordering: DROID-SLAM has lower APE in 14 of 15 settings, with fog and contrast at severity 5 as the only reversal. Figure~\ref{fig:selection} shows the full distribution rather than only its mean.

This result should not be read as a component ablation. The experiments do not isolate the contribution of dense correspondences, recurrent updates, or global optimization. They do show that an image-level proxy is not simply a less accurate version of a geometry-aware proxy. The two can alter correspondence support differently and lead to opposite decisions about the preferred system.

\resultbox{\textbf{Corruption fidelity changes the preferred tracker.} The image-space study mostly favors DPVO, while every paired geometry-aware setting and 14 of 15 compound settings favor DROID-SLAM.}

\subsection{Are the Trends Consistent across Image-Space, Geometry-Aware, and Compound Effects?}
There is no method-independent mapping between the proxy layers. Figure~\ref{fig:transfer}(b) compares geometry-aware APE curves with their compound counterparts. For DPVO, low light and brightness with contrast have identical severity ordering ($\rho=1.00$), while rain and rain with contrast are almost unrelated ($\rho=-0.10$). DROID-SLAM shows the opposite pattern: its rain curves correlate strongly ($\rho=0.80$), whereas its low-illumination curves do not ($\rho=-0.10$). Fog is similarly method dependent, with $\rho=0.90$ for DPVO and $\rho=-0.20$ for DROID-SLAM.

Nominal severity is not a common difficulty coordinate either. Under geometry-aware rain, DROID-SLAM APE decreases from \SI{111.9}{m} at severity 1 to \SI{87.9}{m} at severity 5, while DPVO remains between \SI{125.5}{m} and \SI{129.2}{m}. This decrease does not mean that the tracker improves as rain becomes more severe. The corruption changes which image structure remains usable and, in turn, the trajectory reached by the optimizer. For SLAM, severity orders the corruption parameters, not necessarily the estimation difficulty.

A single mean corruption score or a single correlation can therefore hide the mechanism of interest. Agreement depends on the tracker, the physical effect, the metric, and the portion of the sequence over which the trajectory remains evaluable.

\resultbox{\textbf{No single relationship holds across proxy layers.} Correlations change with the tracker and condition, and nominal corruption severity does not consistently order SLAM difficulty.}

\subsection{Which Conclusions Transfer to Real Adverse Conditions?}
\begin{figure}[t]
  \centering
  \includegraphics[width=\textwidth]{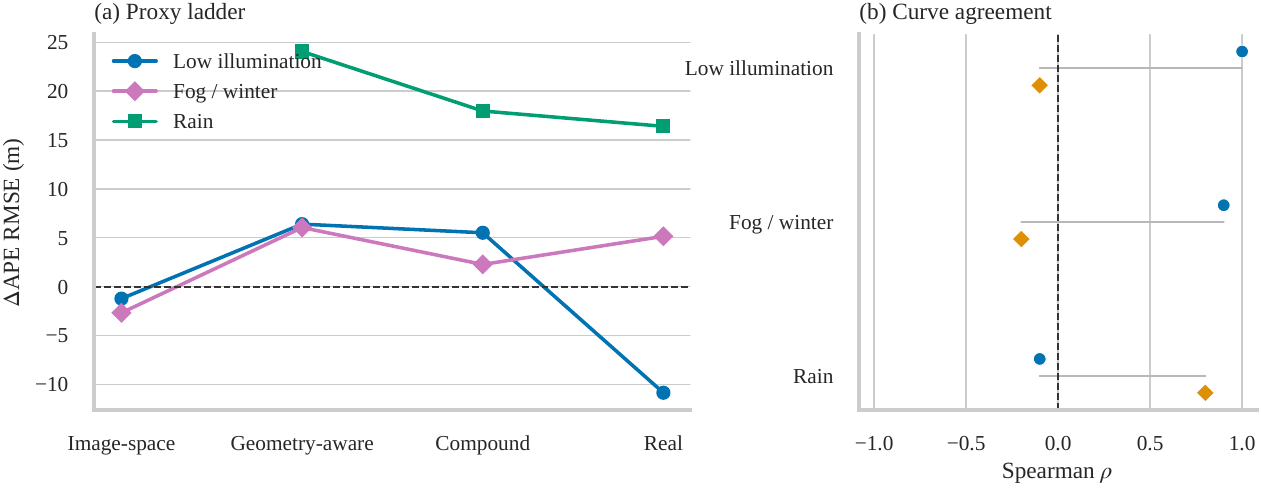}
  \caption{Synthetic-to-real evidence across matched conditions. (a) Learned-tracker margin through the proxy ladder; $\Delta\ape>0$ favors DROID-SLAM. Rain and winter retain the compound-proxy ordering, while real evening illumination reverses to DPVO. (b) Spearman correlation between geometry-aware and compound severity curves; blue circles denote DPVO and orange diamonds denote DROID-SLAM. High agreement between two synthetic curves does not guarantee real-world transfer.}
  \label{fig:transfer}
\end{figure}

Figure~\ref{fig:transfer}(a) follows the learned-tracker margin from a simple image-space transform to a geometry-aware effect, its compound proxy, and the corresponding 4Seasons condition. Rain provides the strongest agreement. DROID-SLAM is better by \SI{17.96}{m} for rain with contrast averaged over severities and by \SI{16.40}{m} on the real rainy sequence. Fog with contrast also selects DROID-SLAM, and the same ordering appears in the real winter sequence, with margins of \SI{2.27}{m} and \SI{5.15}{m}, respectively.

Evening illumination is the counterexample. Brightness with contrast favors DROID-SLAM by \SI{5.51}{m}, but the real cloudy evening favors DPVO by \SI{10.85}{m}. Notably, DPVO has $\rho=1.00$ between the geometry-aware low-light and compound severity curves. Perfect agreement between the two synthetic constructions still fails to preserve the real tracker ordering. Global brightness and contrast do not capture camera-specific noise, automatic exposure, spatially varying illumination, headlights, motion-dependent blur, or the route itself~\cite{stracke2025vision,muller2023classification,fatima2026rawdet}.

The transfer result is therefore two out of three adverse conditions, not a claim of synthetic-real equivalence. Absolute error magnitudes are not comparable across different routes, sensors, and motion profiles. The useful conclusion is narrower: structured rain and fog proxies can reveal an architectural weakness that persists in natural data, while an intuitive illumination proxy can select the wrong tracker.

\resultbox{\textbf{Only some conclusions transfer.} Rain and winter preserve the compound-proxy ordering; evening illumination does not, and none of the synthetic settings predicts absolute real-world error.}

\section{Discussion}
\label{sec:discussion}

The experiments expose not only different levels of robustness, but fundamentally different ways of failing. ORB-SLAM2~\cite{murartal2017orbslam2} uses the same ORB~\cite{rublee2011orb} representation for tracking, local mapping, relocalization, and loop detection. This coupling is both a strength and a weakness. When a sufficient set of repeatable keypoints survives, geometric verification, bundle adjustment, and loop closure can reject inconsistent matches and correct accumulated drift. When noise, blur, or structured occlusion destabilizes that feature set, the corruption simultaneously weakens frame-to-map tracking and the evidence required for relocalization or loop closure. The frequent empty trajectories under noise and snow are therefore consistent with a threshold-like breakdown of the shared feature representation. The route-dependent clean results follow the same reasoning: explicit global correction is valuable when reliable correspondences and revisited regions are available, but it cannot compensate for insufficient or ambiguous visual evidence.

DPVO~\cite{teed2023dpvo} replaces detector-selected keypoints with a sparse set of learned feature patches. A recurrent update operator estimates patch trajectories across nearby frames, predicts a confidence weight for each correspondence factor, and updates patch depths and camera poses through differentiable bundle adjustment. This soft, learned correspondence model provides a plausible mechanism for the observed tracking continuity after the ORB front end has already failed. However, DPVO is a visual odometry system: its online optimization is restricted to a local window and does not include loop closure or a global correction backend. Small but persistent correspondence biases can consequently remain locally compatible while accumulating into substantial global drift.

DROID-SLAM~\cite{teed2021droid} differs from DPVO in two important respects. It reasons over dense feature correlations rather than a sparse patch set, and its backend performs global bundle adjustment over the keyframe history, including long-range graph connections for loop closure. Dense evidence provides redundancy when individual image regions become unreliable, while joint multi-frame refinement can distribute and correct local correspondence errors. This design is consistent with DROID-SLAM obtaining lower APE across the available geometry-aware comparisons and in 14 of the 15 compound settings. Nevertheless, global optimization can only be as reliable as the evidence supplied to it. If a scene-wide corruption systematically biases the dense correspondence field, bundle adjustment may produce a smooth and internally consistent trajectory that remains globally incorrect. Continued output, low local residuals, or low RPE therefore do not by themselves certify low accumulated drift. Since this work does not ablate individual architectural components, these links should be interpreted as mechanism-consistent explanations rather than causal attributions.

This distinction also sharpens the difference between availability and trustworthiness. DPVO~\cite{teed2023dpvo} and DROID-SLAM~\cite{teed2021droid} already predict confidence weights that control how individual correspondence factors contribute to bundle adjustment. These weights are learned indirectly through the pose-estimation objective; they are not supervised as calibrated probabilities that the final trajectory is correct. They should therefore not be interpreted as trajectory-level uncertainty without additional validation. A deployment-facing system should expose signals that are tested against accumulated drift and impending tracking loss, potentially combining correspondence confidence, geometric residuals, graph connectivity, agreement between local and global estimates, and temporal consistency. The relevant question is not only whether the tracker can return another pose, but whether that pose remains sufficiently constrained to be trusted.

The results further show why corruption design must follow the mechanism being tested. Image-space transformations remain useful for broad and inexpensive screening, particularly for compression artifacts, global intensity changes, blur, and high-frequency noise~\cite{agnihotri_unreasonable,agnihotri2024beware,agnihotri2024improving,keuper2013blind}. They primarily probe the stability of the visual front end. Geometry-aware corruptions additionally alter which parts of the scene remain informative as a function of depth and spatial structure, while compound corruptions test whether several individually manageable changes interact inside the correspondence and optimization pipelines. The reversal of the preferred learned tracker across different proxy layers shows that no single image-space corruption score can represent robustness in general. A useful evaluation should instead form a proxy ladder: image-space corruptions for breadth, geometry-aware and temporally coherent corruptions for mechanism fidelity, and natural adverse-condition sequences for external validation. RobustSpring~\cite{oei2026robustspring} provides an instructive example of preserving temporal, stereo, and depth consistency when corrupting correspondence data.

Finally, recent dense systems couple tracking quality to the state of the reconstructed map. Gaussian Splatting SLAM~\cite{matsuki2024gaussian} optimizes camera poses directly against an incrementally updated Gaussian scene representation. An adverse observation can therefore influence both the current pose estimate and the map that will be used to track subsequent frames. MASt3R-SLAM~\cite{murai2025mast3r} combines a learned two-view reconstruction prior with pointmap matching, local fusion, loop closure, and global optimization. SLAM3R~\cite{liu2025slam3r} takes a substantially different route by directly regressing local pointmaps and registering them into a global reconstruction without explicitly solving for camera parameters. These systems may continue producing visually plausible dense output even when geometry becomes incomplete, warped, or inconsistent. Robustness evaluation must consequently distinguish trajectory validity, map validity, and systems validity. In addition to APE and RPE, future evaluations should measure reconstruction accuracy and completeness, cross-view geometric consistency, rendering degradation, map growth, runtime stability, and memory consumption.

\section{Limitations}
\label{sec:limitations}
The controlled sweep uses one KITTI~\cite{geiger2012kitti} trajectory and one archived execution per condition, so scene dependence, geographical diversity~\cite{basu_geodiv}, initialization variance, and runtime nondeterminism are not measured. The binary validity criterion distinguishes evaluable output from missing output but does not quantify how much of a truncated trajectory is supported in time. Several configurations are absent from the archive; all paired analyses use only complete DPVO--DROID-SLAM conditions and expose their sample counts. The 4Seasons~\cite{wenzel2021fourseasons} conditions are semantically matched rather than paired recordings of the same route and camera, so they support comparisons of relative behavior, not causal or numerical equivalence. The study is monocular and trajectory-centered, and does not cover other imaging modalities~\cite{poggi2025smart}, inertial fusion, map accuracy, reconstruction completeness, runtime, memory, or energy.

\section{Future Work}
\label{sec:future}
An extension will introduce temporally coherent, sensor-aware corruptions and a broader set of matched natural conditions, together with map-quality and resource measurements. The immediate aim is to establish which controlled proxies reproduce specific field failure modes, rather than to compress accuracy, continuity, reconstruction, and efficiency into one score.
Structured graphical formulations provide another possible direction~\cite{keuper2011hierarchical,kardoost2018solving}.
Alternative feature representations may provide another analysis direction~\cite{walter2026images}.
Internal representation analysis is another complementary direction~\cite{alshami2025aim}.
Additionally adding synthetic corruptions to more real-world datasets and extend the benchmarking and study to more SLAM methods. 

\section{Conclusion}
\label{sec:conclusion}
This work evaluates monocular SLAM under controlled and natural visual degradation. ORB-SLAM2 frequently fails by returning no evaluable trajectory, while DPVO and DROID-SLAM continue producing poses and instead accumulate drift. These are different operational failures and should be reported separately. More importantly, the apparent ranking of the learned trackers changes with the corruption model: image-space effects mostly favor DPVO, whereas geometry-aware and compound effects strongly favor DROID-SLAM.

Synthetic corruptions are therefore valuable as controlled diagnostic interventions, but not as calibrated forecasts of deployment error. Rain and winter preserve the compound-proxy ordering in real data; evening illumination does not. Robustness evaluation should ask whether tracking remains available, whether its error stays bounded, and whether the conclusion survives the natural condition the proxy is intended to represent. 

\section*{Acknowledgments}
Shashank Agnihotri and Margret Keuper acknowledge support by DFG Research Unit 5336 -- Learning to Sense.
Margret Keuper acknowledges funding by BMFTR project TrackOpt (01IS24074A-D).
The authors acknowledge support by the state of Baden-Württemberg through bwHPC.
The authors acknowledge support by the state of Baden-Württemberg through bwHPC and the German Research Foundation (DFG) through grant INST 35/1597-1 FUGG.
The authors gratefully acknowledge the computing time provided on the high-performance computer HoreKa by the National High-Performance Computing Center at KIT (NHR@KIT). This center is jointly supported by the Federal Ministry of Research, Technology and Space and the Ministry of Science, Research and the Arts of Baden-Württemberg, as part of the National High-Performance Computing (NHR) joint funding program. HoreKa is partly funded by the German Research Foundation (DFG).

\section*{Resources}
The dataset is available at: \url{https://data.dws.informatik.uni-mannheim.de/machinelearning/Failure_or_Drift_dataset/}.

The code is available at: \url{https://github.com/abhaythomas/master_thesis_vslamlab_robustness}.

\bibliographystyle{splncs04}
\bibliography{main}

\end{document}